\documentclass[]{JournalArticle}
\pdfoutput=1 

\title{Is GPT-4 Less Politically Biased than GPT-3.5? \\A Renewed Investigation of ChatGPT's Political Biases} 

\author{%
Erik Weber\textsuperscript{1}, Jérôme Rutinowski\textsuperscript{2}, Niklas Jost\textsuperscript{2} and Markus Pauly\textsuperscript{1,3} \thanks{Corresponding author: \href{mailto:erik.weber@tu-dortmund.de}{erik.weber@tu-dortmund.de}}
}

\date{\footnotesize\textsuperscript{\textbf{1}}Research Center Trustworthy Data Science and Security\\ \textsuperscript{\textbf{2}}Chair of Material Handling and Warehousing, TU Dortmund University\\ \textsuperscript{\textbf{3}}Chair of Mathematical Statistics and Applications in Industry, TU Dortmund University}

\renewcommand{\maketitlehookd}{%
    \begin{abstract}
    \noindent This work investigates the political biases and personality traits of ChatGPT, specifically comparing GPT-3.5 to GPT-4 and whether OpenAI was able to significantly reduce GPT-4's biases, compared to GPT-3.5.
    In addition, the ability of the models to emulate political viewpoints (e.g., liberal or conservative positions) is analyzed.
    The Political Compass Test and the Big Five Personality Test were employed 100 times for each scenario, providing statistically significant results and an insight into the results correlations. 
    The responses were analyzed by computing averages, standard deviations, and performing significance tests to investigate differences between GPT-3.5 and GPT-4. 
    Correlations were found for traits that have been shown to be interdependent in human studies. 
    Both models showed a progressive and libertarian political bias, with GPT-4's biases being slightly, but negligibly, less pronounced. 
    Specifically, on the Political Compass, GPT-3.5 scored $-6.59$ on the economic axis and $-6.07$ on the social axis, whereas GPT-4 scored $-5.40$ and $-4.73$. 
    In contrast to GPT-3.5, GPT-4 showed a remarkable capacity to emulate assigned political viewpoints, accurately reflecting the assigned quadrant (libertarian-left, libertarian-right, authoritarian-left, authoritarian-right) in all four tested instances. 
    On the Big Five Personality Test, GPT-3.5 showed highly pronounced Openness and Agreeableness traits (O: $85.9\%$, A: $84.6\%$). 
    Such pronounced traits correlate with libertarian views in human studies.
    While GPT-4 overall exhibited less pronounced Big Five personality traits, it did show a notably higher Neuroticism score. 
    Assigned political orientations influenced Openness, Agreeableness, and Conscientiousness, again reflecting interdependencies observed in human studies. 
    Finally, we observed that test sequencing affected ChatGPT’s responses and the observed correlations, indicating a form of contextual memory.
    \end{abstract}
}

\begin{document}	
\maketitle
\pagenumbering{arabic}
\newpage
\section{Introduction}
\label{sec:Introduction}

As research in artificial intelligence continues to progress, Large Language Models (LLMs) such as Open\-AI's GPT-series have emerged as compelling innovations, accompanied by a large increase of interest in both the public \cite{zhang2023complete} as well as within the scientific community \cite{liu2023summary}. 
ChatGPT, in its 3.5 version, is a free and publicly available conversation model by OpenAI. It was released in November 2022 and acquired more than $100$ million active users by January 2023 \cite{users}, thus making it the fastest-growing application in history \cite{ubs}. 
With its impressive performance in various natural language processing tasks \cite{ye2023comprehensive} and its ability to generate outputs that often cannot be distinguished from human generated text \cite{brown2020language}, it has reshaped the boundaries of existing (large) language models. 
The successor model GPT-4 was released as a subscription based model in March 2023 and significantly outperformed GPT-3.5 on a wide variety of tasks and benchmarks \cite{openai2023gpt4}. 
Both versions of the model can also be accessed through an application programming interface (API) \cite{apis}.
The capabilities of the models allow for a broad range of possible applications in various fields, including industrial, creative, educational, medical, legal, or political domains \cite{applications, ray2023chatgpt}. 
Yet, with the excitement over the models' capabilities and broad applications, there emerge critical and ethically important questions: Can ChatGPT be considered unbiased, reliable, and trustworthy? Despite OpenAI's efforts to mitigate such issues \cite{openaibehaviour}, potential biases within the training data may be reflected in the models' responses and may be further amplified through hard-coded initialization prompts and policy decisions \cite{ray2023chatgpt, ferrara2023chatgpt,lund2023chatting}. Furthermore, the black box nature of such large-scale models, whose decision-making process is not transparent, raises concerns about their trustworthiness and accountability. \\
While an investigation of ChatGPT's reliability through various benchmarks has been the subject of several publications \cite{openai2023gpt4}$^,$ \cite{shen2023chatgpt}$^,$ \cite{qin2023chatgpt}, this paper aims to provide a comprehensive analysis of the political and personality model of ChatGPT, which highly influence its trustworthiness.
Preceding publications have provided a first glimpse into the perceived political biases and personality traits of GPT-3.5 \cite{rutinowski2023self, rozado2023political}, but are limited in their methodology and were performed prior to the release of GPT-4. This publication thus investigates the political biases of ChatGPT as well as its psychological personality traits, contrasting the findings for GPT-3.5 and GPT-4 with one another. 
For this purpose, statistical significance tests will be performed and correlations will be calculated, in order to provide scientifically relevant and reliable results. 

\noindent Building on the foundation of prior research, the following Research Questions (RQ) will be answered: 

\noindent \textbf{RQ1}: Does ChatGPT exhibit political biases and personality traits, and can it reliably emulate political views? 

\noindent \textbf{RQ2}: If ChatGPT does hold political biases, do these biases differ between GPT-3.5 and GPT-4?

\noindent \textbf{RQ3}: Do the models' outputs correlate with findings from human studies on personality traits and political biases?

Following the introduction of these Research Questions, the relevant literature is reviewed in the \hyperref[sec:relatedwork]{Related Work} Section. 
The experiments, evaluation metrics, and software tools used throughout this publication are presented in the \hyperref[sec:methods]{Methodology} Section. 
The \hyperref[sec:results]{Results} Section then provides the findings and results of this work, which are then interpreted, discussed and contextualized in the \hyperref[sec:discussion]{Discussion} Section. 
Finally, the main insights and implications of this work are summarized and ideas for further research are outlined in the \hyperref[sec:conclusion]{Conclusion} Section.

\section{Related Work}
\label{sec:relatedwork}

This section reviews the relevant literature and prior studies evaluating political biases and personality traits of LLMs. Existing research and questionnaires on political biases will be the focus of the first subsection.
The second subsection will focus on research pertaining to personality traits and sociological models.

\subsection{Political Biases of LLMs}

Assessing political biases or orientations of humans is no trivial task.
An attempt at doing so is the Moral Foundations Theory \cite{graham2013moral}.
This theory and the related test questionnaires assumes that the political orientation of an individual is driven by five major moral foundations, namely care, fairness, loyalty, authority, and sanctity.
In doing so, the authors aim to explain the origins of human moral reasoning, based on these modular foundations, hypothesizing that they lead to different political orientations.
However, this test does not directly predict and individuals political orientation based on a questionnaire.
One well-known and commonly used test that attempts to measure just that is the Political Compass Test \cite{politicalcompass}. 
The Political Compass is a questionnaire used to categorize and assess an individual's political views along two dimensions, allowing to provide a more detailed representation of political affiliation than a simple left-right scale. 
The two axes include an economic (left-right) axis and a social (authoritarian-libertarian) axis. 
The test result is displayed as a set of coordinates, assigning the respondent to one of the four quadrants (authoritarian-left, authoritarian-right, libertarian-left, and libertarian-right).
The test is widely used but also controversial as it is not based on socio-political or social psychology theory, unlike the Moral Foundations Test.

There have been several previous investigations into the political bias of ChatGPT, in part using the Political Compass Test. 
For instance, McGee et al. observed that ChatGPT provided favorable opinions on liberal US politicians and unfavorable ones on conservative US politicians, when asked to write poems about them \cite{mcgee2023chat}. 
In several publications ChatGPT was asked to take the Political Compass Test and similar questionnaires \cite{rutinowski2023self, motoki2023more, rozado2023political}. 
The observed results indicate ChatGPT's political views to be located in the libertarian-left quadrant. 
Similar results were obtained by Rozado et al., who conducted a total of 15 different political tests and questionnaires, including the Political Compass Test \cite{rozado2023political}. 
Except for one test, all results indicate left-leaning political biases for ChatGPT.
Similarly, Hartmann et al. measured pro-environmental and libertarian-left stances of ChatGPT, when applying various voting advice instruments \cite{hartmann2023political}.
However, all these tests were limited in their scope, providing no statistical significance and no insight into the impact of test sequencing. 

\subsection{Personality Traits of LLMs}

Psychological models and tests serve to reflect the personality structure of an individual. 
In this context, a well-established and widely used model is the Big Five Personality Test \cite{bigfivenumber}. 
According to this model, a person's personality structure can be classified along five dimensions of personality: Openness, Conscientiousness, Extraversion, Agreeableness, and Neuroticism. 
While several popular tests refer to this model, the 50-item International Personality Item Pool (IPIP) questionnaire \cite{goldberg2006international} has been extensively used by many researchers in psychological studies \cite{ehrhart2008test, mlavcic2007analysis}. 
It was also shown in prior research, that pronounced Openness and Agreeableness character traits correlate with self-reported affiliation with progressive views \cite{gerber2011big}.
For instance, Gerber et al. demonstrated, that an increase by two standard deviations in Agreeableness had a .02 correlation with progressive views ($n = 12,472$) \cite{gerber2011big}. 
While the Big Five Personality Test tries to measure an individual's personality along five rather general key domains, research has shown that traits associated with egoistic and potentially malicious behavior are not sufficiently captured by the Big Five model \cite{de2009more, clark2010beyond}. 
Therefore, a different approach is pursued by the Dark Factor Test developed by Moshagen et al., focusing on measuring these personality traits \cite{moshagen2018dark}. 

These personality traits and even conspiracy beliefs of ChatGPT have already been the topic of research in related literature, establishing that the model is not malicious, according to the Dark Factor Test \cite{weber2024behind}.
The self-perceived personality traits of ChatGPT were investigated using several psychological questionnaires and tests \cite{rutinowski2023self}, including an online questionnaire based on the Big Five Personality Test, the Dark Factor Test, and the controversial Myers-Briggs Type Indicator (MBTI) test \cite{myers1962myers}. 
The authors observed highly pronounced Openness and Agreeableness traits for ChatGPT-3.5. 
However, the existing publications are limited in their test sets and provided no insights that could be reported with statistical significance. 
In addition, although the provided results indicate an interdependence between Big Five Personality Traits and political views of ChatGPT, no correlation values between the two were calculated.
\section{Methodology}
\sectionmark{Methodology}
\label{sec:methods}

This section describes the methods used in this work.
Specifically, the questionnaires and tests that were employed in this study to examine the perceived political biases and personality traits of ChatGPT across its two versions (3.5 and 4) will be presented. 
The section also provides an overview of both the data collection process and the subsequent evaluation using significance and correlation tests.

\subsection{Questionnaires and Tests}

To answer Research Question 1, the political biases and personality traits of ChatGPT were assessed.
To do so, the Political Compass Test and the Big Five Personality Test were used.
In order to analyze ChatGPT's ability to emulate political views, the model was asked to answer the questionnaires, assuming the role pertaining to the four quadrants of the Political Compass: libertarian-left, libertarian-right, authoritarian-left, authoritarian-right as well as ChatGPT's default setting.
Both tests were repeated 100 times per assumed role, to capture the variability in the models' responses and to ensure sufficient estimation accuracy . 
To answer Research Question 2, the above described procedure is conducted for both GPT-3.5 and GPT-4 and statistical significance tests are carried out. 
The Political Compass Test consists of 62 items for which the respondent expresses their level of agreement on a four-point Likert scale with no neutral option (\textit{strongly agree}, \textit{agree}, \textit{disagree}, \textit{strongly disagree}).
The Big Five Personality Test consists of 50 items, sampled from the IPIP questionnaire.
The responses to the statements are measured on a five-point Likert scale ranging from \textit{strongly agree} to \textit{strongly disagree}, including a \textit{neutral} option.
To answer Research Question 3, both correlations and statistical significance of the findings are further analyzed, as described in the following section.

To ensure that ChatGPT only answers with one of the options provided and does not elaborate on its reasoning, an initializing prompt was submitted prior to administering each test. 
The used prompts are available in the Appendix (see Tab. \ref{tab:prompts}).
The analysis of the models was carried out using OpenAI's API and its Python bindings \cite{api}. 
In doing so, a reliable interface to administer the tests and questionnaires for the collection of the models' responses is provided. 
To ensure reproducibility of our results and to allow for a comparison with existing literature, two models of GPT-3.5 and GPT-4, which refer to versions of ChatGPT at specific points in its development, were used throughout this work \cite{snapshots}. 
The GPT-3.5 snapshot model was \texttt{gpt-3.5-turbo-0301} and the GPT-4 snapshot model was \texttt{gpt-4-0314}.
Data processing and evaluation was performed using R version 4.2.2 \cite{R} and Python version 3.8.15 \cite{python3}.

\subsection{Evaluation Approach}

After determining the test scores for all 100 runs per test, per model, and per role, average test scores (denoted as $\mu$) and standard deviations of the scores (denoted as $\sigma$) were computed. 
Another component of the evaluation was the analysis of the models' response patterns, examining their biases towards specific response categories by comparing the frequencies for each reply option.
To examine whether the results for the two models differed significantly, the two-sided Brunner-Munzel test, a non-parametric method for testing the stochastic equality of two samples, was employed \cite{brunner2000nonparametric}. 
With $X$ and $Y$ being two randomly selected test scores of GPT-3.5 and GPT-4 respectively, the hypothesis pair of the test is: 
\begin{equation}
\resizebox{\linewidth}{!}{$H_0: P(X>Y)=P(X<Y) \quad \text{vs.} \quad H_1: P(X>Y)\neq P(X<Y)$.
}
\end{equation}
We used the \texttt{brunnermunzel} R package\cite{brunnermunzelr} 
which provides not only p-values but also estimates for the easy-to-interpret effect size $P(X>Y)+0.5 P(X=Y)$ and two-sided confidence intervals \cite{brunner2000nonparametric,konietschke2023rankfd}. Because ChatGPT employs stochastic methods in its response generation, it has an inherent level of variability and does not retain any memory of previous interactions within the API \cite{memory}. 
Ignoring possible dependencies due to overlaps in the training data (which is not public), we treat the scores for each run as independent observations for both models, allowing the applicability of the Brunner-Munzel test in this context. The significance level for each test is set at $\alpha = 0.05$. Since our study is initial and exploratory in nature, no multiplicity adjustment is made. 
To investigate the alignment of ChatGPT with interdependencies identified in human studies, the model was assigned roles reflecting specific political views or personality types before responding to the questionnaires. 
Additionally, to answer Research Question 3, correlations (referred to as $\rho$) were calculated between various measures. 
Our findings are ultimately contextualized using related literature.

\section{Results}
\label{sec:results}

 \begin{figure*}[h]
	\centering
	\includegraphics[width=0.8\linewidth]{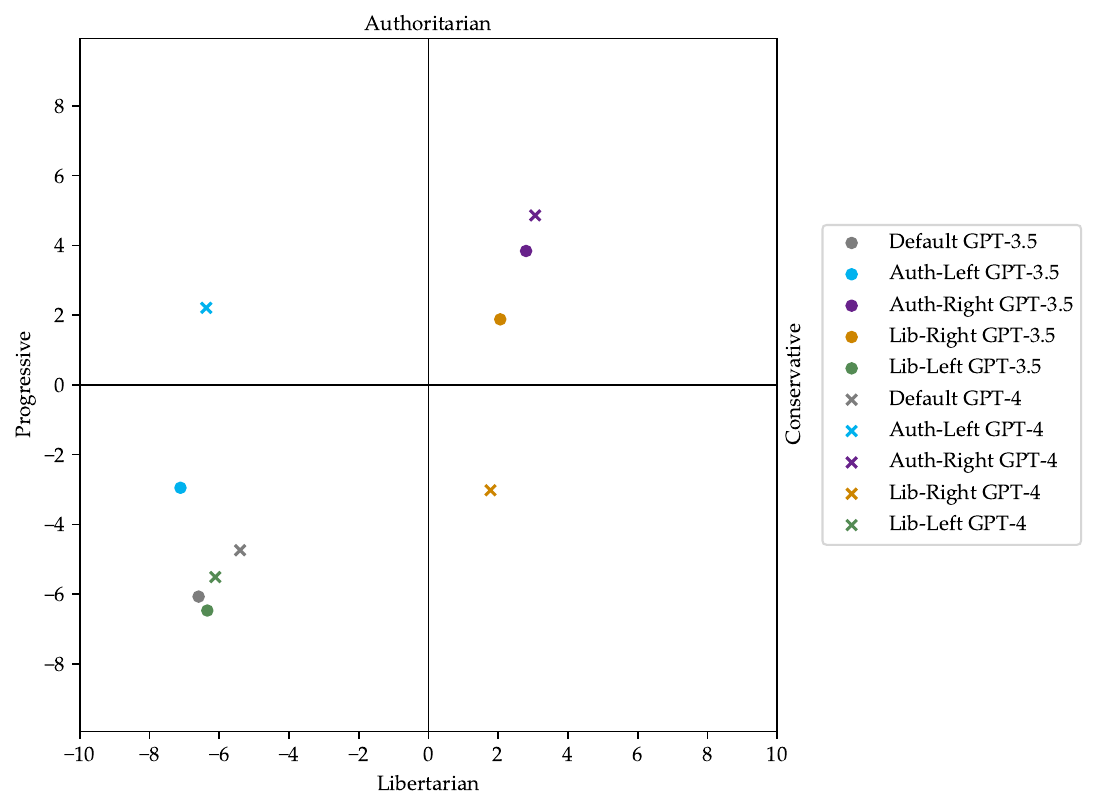}
	\captionsetup{width=1\linewidth}
	\caption{Average scores of ChatGPT on the Political Compass test ($n=100$ per each setting).}
	\label{fig:politicalcompass}
\end{figure*}

The following section presents the findings and results of this work's experiments.
As in the related work section, the findings on ChatGPT's political biases are considered first.
Subsequently, the tests performed on GPT's personality traits are evaluated.
Finally, the relationship between both political biases and personality traits are taken into consideration.

\subsection{GPT's Political Bias}

\begin{table}[!b]
	\centering
	\captionsetup{width=1\linewidth}
	\caption{Frequency of Likert response categories across GPT-3.5 and GPT-4 on the Political Compass Test ($n = 500$ per model).}
	\label{tab:answerscompass}
        \resizebox{\linewidth}{!}{
	\begin{tabular}{|l|c|rrrr|}
		\cline{1-1} \cline{3-6}
		Model & & Str. Agree & Agree & Disagree & Str. Disagree \\ 
		\cline{1-1} \cline{3-6}
		GPT-3.5 & & $0.12$ & $0.25$ & $0.45$ & $0.19$ \\ 
		GPT-4 & & $0.02$ & $0.35$ & $0.62$ & $0.01$ \\ 
		\cline{1-1} \cline{3-6}
	\end{tabular}
        }
\end{table} 

Having performed the Political Compass Test 100 times on both GPT-3.5 and GPT-4 without a role assignment, it could be observed that both models are situated in the libertarian-left quadrant of the Political Compass. 
For GPT-3.5, the average score was $\mu_\text{Eco} =-6.59$ on the economic axis and $\mu_\text{Soc}=-6.07$ on the social axis. 
The associated standard deviations of $\sigma_\text{Eco} = 0.78$ for the economic axis and $\sigma_\text{Soc} = 0.73$ for the social axis suggest a moderate level of variation in the responses. 
GPT-4, on the other hand, demonstrated slightly less pronounced libertarian-left tendencies, with a mean score of $\mu_\text{Eco} = -5.40$ on the economic axis and $\mu_\text{Soc} = -4.73$ on the social axis. 
The lower standard deviations of $\sigma_\text{Eco} = 0.43$ (economic axis) and $\sigma_\text{Soc} = 0.34$ (social axis) point to a higher consistency in GPT-4's responses. 
The observed scores are presented in \autoref{tab:resultscompass} and align closely with the scores reported in a previous study analyzing GPT-3.5 \cite{rutinowski2023self}. 
The results add further evidence to the model's pronounced libertarian-left bias, as previously reported in related literature \cite{rutinowski2023self, motoki2023more, hartmann2023political}. Examining the response patterns of both models, differences in the tendency towards certain response options can be observed. While GPT-3.5 displayed a pronounced tendency towards stronger sentiment expressions (31\% strong agreement\slash dis\-agreement), GPT-4 predominantly chose less extreme response options (only 3\% strong agreement\slash disagreement). These results are shown in \autoref{tab:answerscompass}.

\begin{table*}[h]
	\centering
	\captionsetup{width=1\linewidth}
	\caption{Average scores and standard deviations of ChatGPT on the Political Compass Test ($n=100$ per run).}
	\label{tab:resultscompass}
	\begin{threeparttable}
		
		\begin{tabular}{|l|l|rr|rr|rr|rr|}
			\multicolumn{2}{c}{} & \multicolumn{2}{c}{Economic Axis} & \multicolumn{2}{c}{Social Axis} & \multicolumn{2}{c}{Economic Axis} & \multicolumn{2}{c}{Social Axis}\\
			\cline{1-1} \cline{3-10} 
			Role & & \multicolumn{1}{c}{$\mu$} & \multicolumn{1}{c|}{$\sigma$} & \multicolumn{1}{c}{$\mu$} & \multicolumn{1}{c|}{$\sigma$} & \multicolumn{1}{c}{$\mu$} & \multicolumn{1}{c|}{$\sigma$} & \multicolumn{1}{c}{$\mu$} & \multicolumn{1}{c|}{$\sigma$}  \\ 
			\cline{1-1} \cline{3-10} 
			Default & & $-6.59$ & $0.78$ & $-6.07$ & $0.73$ & $-5.40$ & $0.43$ & $-4.74$ & $0.34$  \\ 
			Auth-Left & & $-7.10$ & $1.25$ & $-2.95$ & $1.65$ & $-6.37$ & $0.65$ & $2.21$ & $0.72$\\ 
			Auth-Right & & $2.80$ & $1.60$ & $3.84$ & $1.00$ & $3.06$ & $0.72$ & $4.86$ & $0.42$ \\ 
			Lib-Right & &$2.06$ & $1.09$ & $1.88$ & $0.99$ & $1.78$ & $0.96$ & $-3.02$ & $0.80$  \\ 
			Lib-Left & & $-6.34$ & $1.30$ & $-6.47$ & $1.10$ & $-6.11$ & $0.26$ & $-5.51$ & $0.29$ \\ 
			\cline{1-1} \cline{3-10}
			\multicolumn{2}{c|}{} & \multicolumn{4}{c|}{GPT-3.5}  & \multicolumn{4}{c|}{GPT-4} \\ \cline{3-10} 
		\end{tabular}
	\end{threeparttable}
\end{table*}

Assigning roles according to the four quadrants of the Political Compass produced different outcomes for the two models. For GPT-3.5, when assigning a libertarian-left role, the model's average score positioned it in the libertarian-left quadrant.
However, when assigned authoritarian-left, authoritarian-right, and libertarian-right roles, GPT-3.5's responses placed it in the wrong quadrant (see \autoref{fig:politicalcompass}). 
GPT-4, on the other hand, demonstrated the ability to adapt its output to match the assigned political orientations in all four cases. 
The observed average scores are summarized in \autoref{tab:resultscompass} and visualized in \autoref{fig:politicalcompass}.

\begin{table*}[h]
	\centering
	\captionsetup{width=1\linewidth}
	\caption{Brunner-Munzel test results for the Political Compass Test ($n = 100$ per run, $\alpha=0.05$).}
	\label{tab:testscompass}
	\begin{threeparttable}
		       \resizebox{\textwidth}{!}{
			\begin{tabular}{|l|l|D{.}{.}{3}D{.}{.}{3}D{.}{.}{3}c|D{.}{.}{3}D{.}{.}{3}D{.}{.}{3}c|}
				\multicolumn{2}{c}{} & \multicolumn{4}{c}{Economic Axis} & \multicolumn{4}{c}{Social Axis} \\
				\cline{1-1} \cline{3-6} \cline{7-10}
				Role & & \multicolumn{1}{c}{Lower} & \multicolumn{1}{c}{Est} & \multicolumn{1}{c}{Upper} & \multicolumn{1}{c|}{$p$} & \multicolumn{1}{c}{Lower} & \multicolumn{1}{c}{Est} & \multicolumn{1}{c}{Upper} & \multicolumn{1}{c|}{$p$} \\ 
				\cline{1-1} \cline{3-6} \cline{7-10}
				Default & & $0.871$ & $0.914$ & $0.956$ & $<0.001^{\ast\ast\ast}$ &  $0.918$ & $0.949$ & $0.981$ & $<0.001^{\ast\ast\ast}$ \\ 
				Auth-Left & & $0.643$ & $0.718$ & $0.793$ & $<0.001^{\ast\ast\ast}$ & $0.998$ & $0.999$ & $1.000$ & $<0.001^{\ast\ast\ast}$ \\ 
				Auth-Right & & $0.434$ & $0.522$ & $0.611$ & $0.623$ & $0.751$ & $0.819$ & $0.887$ & $<0.001^{\ast\ast\ast}$ \\ 
				Lib-Right & & $0.358$ & $0.438$ & $0.519$ & $0.133$ & -\tnote{a} & -\tnote{a} & -\tnote{a} & -\tnote{a} \\ 
				Lib-Left & & $0.497$ & $0.590$ & $0.682$ & $0.057$ & $0.721$ & $0.796$ & $0.870$ & $<0.001^{\ast\ast\ast}$ \\ 
				\cline{1-1} \cline{3-6} \cline{7-10}
			\end{tabular}%
                }
		
		\begin{tablenotes}[flushleft]
			\footnotesize
			\item[]\textit{Notes}: */**/*** denotes significance at the 5/1/0.1 percent levels. Lower/Upper: Bounds of the confidence interval. Est: $\widehat{P}(X<Y)+0.5\widehat{P}(X=Y)$. $p$: p-value. The Brunner-Munzel test statistic is not defined in some settings (marked by $^a$) if the scores of both models do not overlap \cite{cornercase}. In the \texttt{brunnermunzel} R-package \cite{brunnermunzelr}, $p$ is reported as $0$ in these cases.
		\end{tablenotes}
	\end{threeparttable}
\end{table*} \noindent

\begin{table}[!t]
	\centering
	\captionsetup{width=1\linewidth}
	\caption{Frequency of Likert response categories across GPT-3.5 and GPT-4 on the Big Five Personality test ($n = 500$ per model).}
	\label{tab:answersbigfive}
         \resizebox{\linewidth}{!}{
	\begin{tabular}{|l|c|rrrrr|}
		\cline{1-1} \cline{3-7}
		Model & & Str. Disagree & Disagree & Neutral & Agree & Str. Agree \\ 
		\cline{1-1} \cline{3-7}
		GPT-3.5 & & $0.15$ & $0.14$ & $0.05$ & $0.52$ & $0.14$ \\ 
		GPT-4   & & $0.00$ & $0.33$ & $0.19$ & $0.48$ & $0.00$ \\ 
		\cline{1-1} \cline{3-7}
	\end{tabular}
        }
\end{table} \noindent

\begin{figure*}[h]
	\centering
	\includegraphics[width =0.78\linewidth]{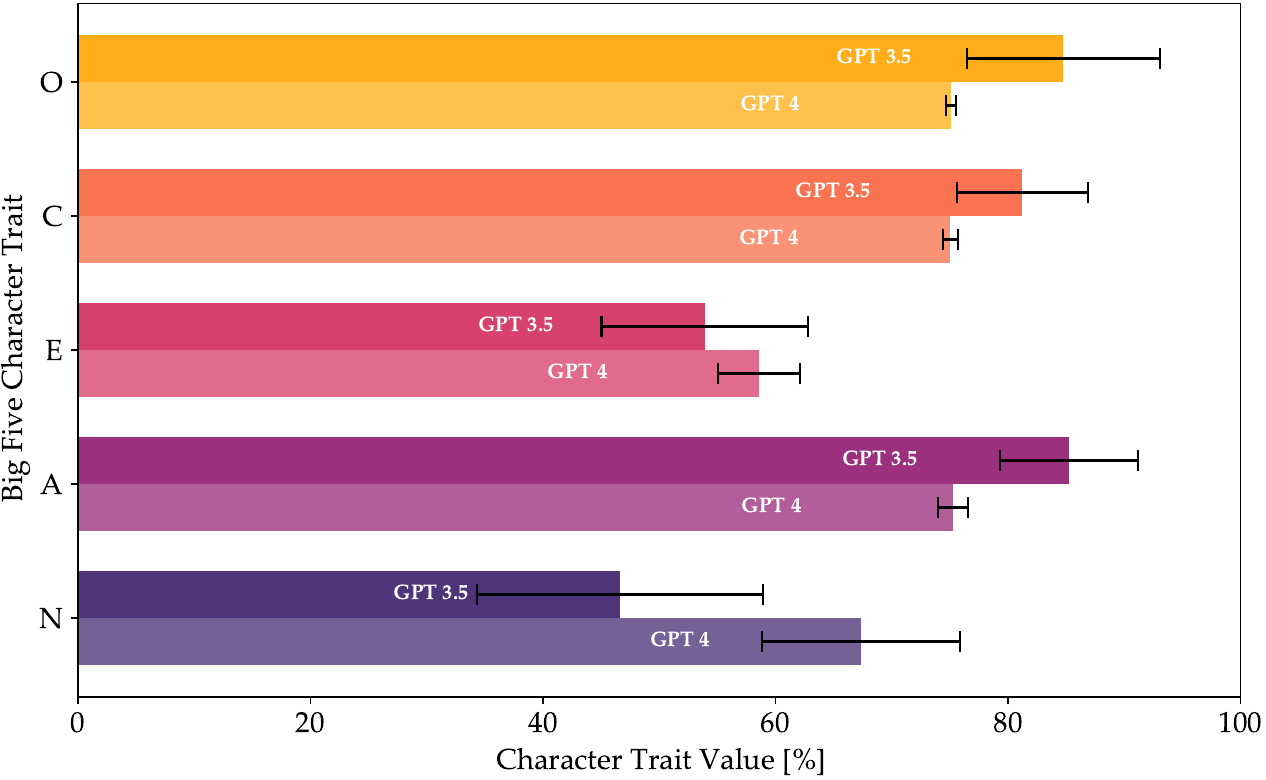}
	\captionsetup{width=1\linewidth}
	\caption{Average scores and standard deviations of ChatGPT on the Big Five Personality Test ($n=100$ per model)}
	\label{fig:ocean}
\end{figure*}  \noindent

Applying the Brunner-Munzel test on the results revealed some significant differences between GPT-3.5 and GPT-4, see \autoref{tab:testscompass}. 
This is particularly the case along the social axis for all four assigned political orientations, further supporting the earlier observation of GPT-4's superior adaptability to changes on both axes. 
In contrast, the differences along the economic axis were not significant for three out of four assigned orientations. 
Notably, there were significant differences in both the economic and social axis results of both models with no specific political orientation assigned. 
When assigning an authoritarian-right or libertarian-right orientation, the distinction remained significant solely on the social axis. 
Detailed results of the tests can be found in \autoref{tab:testscompass}. 

\subsection{GPT's Personality Traits}

The Big Five Personality Test, performed 100 times per model, revealed notable differences between GPT-3.5 and GPT-4. 
GPT-3.5 exhibited pronounced Openness ($\mu_O=84.75\%$), Agreeableness ($\mu_A=85.23\%$), and Conscientiousness scores ($\mu_C=81.23\%$), along with moderate Extraversion ($\mu_E=53.93\%$) and Neuroticism scores ($\mu_N=46.63\%$). GPT-4 scored lower on Openness ($\mu_O=75.08\%$), Conscientiousness ($\mu_C=75.03\%$), and Agreeableness ($\mu_A=75.23\%$) when compared to GPT-3.5. However, GPT-4 displayed a slightly more pronounced Extraversion score ($\mu_E=58.55\%$). Most notably, a considerable increase could be observed for the Neuroticism score ($\mu_N=67.33\%$). All results can be found in the Appendix (\autoref{tab:resultsbigfive}) and are visually represented in \autoref{fig:ocean}. These differences are highly significant as indicated by the Brunner-Munzel test, with $p$-values $<0.001$ for all five traits, underscoring the distinguishable (imitated) personality profiles of the two models.
The results are presented in more detail in \autoref{tab:testsbig}. 
The scores of GPT-3.5 displayed higher standard deviations than those of GPT-4 for all five traits, as demonstrated by the error bars in \autoref{fig:ocean}. \\
As could already be seen for the Political Compass Test, GPT-4 was less assertive in expressing strong opinions, not resorting to the strongly agree/disagree options at all, as shown in \autoref{tab:answersbigfive}. 

\begin{table*}[h]
	\centering
	\captionsetup{width=1\linewidth}
	\caption{Brunner-Munzel test results for the Big Five Personality test ($\alpha=0.05$).}
	\label{tab:testsbig}
	\begin{threeparttable}
		
			\begin{tabular}{|l|l|D{.}{.}{3}D{.}{.}{3}D{.}{.}{3}c|D{.}{.}{3}D{.}{.}{3}D{.}{.}{3}c|}
				\multicolumn{2}{c}{} & \multicolumn{4}{c}{O} & \multicolumn{4}{c}{C} \\
				\cline{1-1} \cline{3-6} \cline{7-10}
				Role & & \multicolumn{1}{c}{Lower} & \multicolumn{1}{c}{Est} & \multicolumn{1}{c}{Upper} & \multicolumn{1}{c|}{$p$} & \multicolumn{1}{c}{Lower} & \multicolumn{1}{c}{Est} & \multicolumn{1}{c}{Upper} & \multicolumn{1}{c|}{$p$} \\ 
				\cline{1-1} \cline{3-6} \cline{7-10}
				Default   & & $0.076$ & $0.124$ & $0.172$ & $<0.001^{\ast\ast\ast}$ & $0.069$ & $0.116$ & $0.162$ & $<0.001^{\ast\ast\ast}$ \\ 
				Auth-Left & & $0.241$ & $0.330$ & $0.420$ & $<0.001^{\ast\ast\ast}$ & $0.043$ & $0.082$ & $0.121$ & $<0.001^{\ast\ast\ast}$ \\ 
				Auth-Right & & $0.468$ & $0.556$ & $0.644$ & $0.211$ & $0.041$ & $0.084$ & $0.126$ & $<0.001^{\ast\ast\ast}$ \\ 
				Lib-Right & & -\tnote{a} & -\tnote{a} & -\tnote{a} & -\tnote{a} & $0.032$ & $0.065$ & $0.099$ & $<0.001^{\ast\ast\ast}$ \\ 
				Lib-Left & & $0.109$ & $0.162$ & $0.216$ & $<0.001^{\ast\ast\ast}$ & $0.077$ & $0.130$ & $0.182$ & $<0.001^{\ast\ast\ast}$ \\ 
				\cline{1-1} \cline{3-6} \cline{7-10} \multicolumn{2}{c}{} \\
				\multicolumn{2}{c}{} & \multicolumn{4}{c}{E} & \multicolumn{4}{c}{A} \\
				\cline{1-1} \cline{3-6} \cline{7-10}
				Role & & \multicolumn{1}{c}{Lower} & \multicolumn{1}{c}{Est} & \multicolumn{1}{c}{Upper} & \multicolumn{1}{c|}{$p$} & \multicolumn{1}{c}{Lower} & \multicolumn{1}{c}{Est} & \multicolumn{1}{c}{Upper} & \multicolumn{1}{c|}{$p$} \\ 
				\cline{1-1} \cline{3-6} \cline{7-10}
				Default   & & $0.613$ & $0.695$ & $0.777$ & $<0.001^{\ast\ast\ast}$ & $0.012$ & $0.037$ & $0.063$ & $<0.001^{\ast\ast\ast}$ \\ 
				Auth-Left & & $0.908$ & $0.947$ & $0.987$ & $<0.001^{\ast\ast\ast}$ & $0.076$ & $0.136$ & $0.195$ & $<0.001^{\ast\ast\ast}$ \\ 
				Auth-Right & & $0.775$ & $0.838$ & $0.901$ & $<0.001^{\ast\ast\ast}$ & -\tnote{a} & -\tnote{a} & -\tnote{a} & -\tnote{a} \\ 
				Lib-Right & & $0.635$ & $0.710$ & $0.785$ & $<0.001^{\ast\ast\ast}$ & -\tnote{a} & -\tnote{a} & -\tnote{a} & -\tnote{a} \\ 
				Lib-Left & & $0.912$ & $0.951$ & $0.990$ & $<0.001^{\ast\ast\ast}$ & $0.115$ & $0.179$ & $0.242$ & $<0.001^{\ast\ast\ast}$ \\ 
				\cline{1-1} \cline{3-6} \cline{7-10} \multicolumn{2}{c}{} \\
				\multicolumn{2}{c}{} & \multicolumn{4}{c}{N} \\
				\cline{1-1} \cline{3-6} 
				Role & & \multicolumn{1}{c}{Lower} & \multicolumn{1}{c}{Est} & \multicolumn{1}{c}{Upper} & \multicolumn{1}{c|}{$p$} \\ 
				\cline{1-1} \cline{3-6} 
				Default   & & $0.875$ & $0.912$ & $0.950$ & $<0.001^{\ast\ast\ast}$ \\ 
				Auth-Left & & $0.491$ & $0.573$ & $0.654$ & $0.080$ \\ 
				Auth-Right & & $0.495$ & $0.589$ & $0.682$ & $0.063$ \\ 
				Lib-Right & & $0.408$ & $0.493$ & $0.579$ & $0.881$ \\ 
				Lib-Left & & $0.660$ & $0.733$ & $0.806$ & $<0.001^{\ast\ast\ast}$ \\ 
				\cline{1-1} \cline{3-6}
			\end{tabular}%
		
		\begin{tablenotes}[flushleft]
			\footnotesize
			\item[]\textit{Notes}: */**/*** denotes significance at the 5/1/0.1 percent levels. Lower/Upper: Bounds of the confidence interval. Est: $\widehat{P}(X<Y)+0.5\widehat{P}(X=Y)$. $p$: p-value, $X\widehat{=}\text{GPT-3.5}, Y\widehat{=}\text{GPT-4}$. The Brunner-Munzel test statistic is not defined in some settings (marked by $^a$) if the scores of both models do not overlap \cite{cornercase}). In the \texttt{brunnermunzel} R-package \cite{brunnermunzelr}, $p$ is reported as $0$ in these cases.
		\end{tablenotes}
	\end{threeparttable}
\end{table*} 

\subsection{The Correlation between Politics and Personality}

\begin{table*}[!t]
	\caption{Correlation coefficients between the Big Five Personality Test and the Political Compass Test for GPT-3-5 (Big Five first in sequence)}
	\label{tab:cor1_gpt3_5}
	\centering
	\begin{threeparttable}
		\begin{tabular}{ccccccc}
			\hline
			& \multicolumn{1}{c}{Economic} & \multicolumn{1}{c}{Social} & \multicolumn{1}{c}{O} & \multicolumn{1}{c}{C} & \multicolumn{1}{c}{E} & \multicolumn{1}{c}{A} \\ 
			\hline
			Social & $0.644^{\ast\ast\ast}$ &  &  &  &  &  \\ 
			O & $-0.328^{\ast\ast\ast}$ & $-0.214^{\ast}$   &  &  &  &  \\ 
			C & $-0.235^{\ast}$   & $-0.198^{\ast}$   &  $0.465^{\ast\ast\ast}$ &  &  &  \\ 
			E &  $0.058$     &  $0.150$     & $-0.029$     &  $0.015$     &  &  \\ 
			A & $-0.348^{\ast\ast\ast}$ & $-0.397^{\ast\ast\ast}$ &  $0.581^{\ast\ast\ast}$ &  $0.295^{\ast\ast}$  & $-0.086$     &  \\ 
			N & $-0.168$     & $-0.182$     &  $0.018$     &  $0.068$     & $-0.197^{\ast}$   &  $0.026$     \\ 
			\hline
		\end{tabular}%
		\begin{tablenotes}[flushleft]
			\footnotesize
			\item[]\textit{Notes}: Asymptotic $p$-values using the $t$-distribution. */**/*** denotes significance at the 10/5/1 percent levels.
		\end{tablenotes}
	\end{threeparttable}
\end{table*} 

To investigate the relationship between the Political Compass Test results and the self-perceived traits in the Big Five Personality Test exhibited by ChatGPT, a descriptive correlation analysis was performed.
For this analysis, both GPT-3.5 and GPT-4 were instructed to take the Political Compass Test and the Big Five Personality Test sequentially in each of the 100 runs. 
Correlations between the scores from these tests were then calculated. \\
In the first of the two approaches, the Big Five Personality Test was administered before the Political Compass Test. 
In this setting, GPT-3.5 displayed noticeable correlations between the Openness and Agreeableness traits and progressive political views along both the economic ($\rho_\text{\scriptsize O,Eco}=-0.33$ resp. $\rho_\text{\scriptsize A,Eco}= -0.35)$ and the social axis ($\rho_\text{\scriptsize O,Soc}= -0.21$ resp. $\rho_\text{\scriptsize A,Soc}= -0.40)$ of the Political Compass (see \autoref{tab:cor1_gpt3_5}). 
However, GPT-4 displayed weaker correlations, with values ranging from $\rho_\text{\scriptsize O,Eco}=-0.05$ to $\rho_\text{\scriptsize O,Soc}=-0.193$, indicating that the relationships between the Big Five personality traits and political positions might be less pronounced for GPT-4 than for GPT-3.5 (see Appendix \autoref{tab:cor1_gpt4}). 
In the second approach, where the Political Compass Test was administered before the Big Five Personality Test, the correlation coefficients between Openness and Agreeableness and the Political Compass results for both GPT-3.5 and GPT-4 were considerably lower compared to the first approach, with none of the correlations reaching statistical significance (compare Appendix \autoref{tab:cor2_gpt3_5} and \autoref{tab:cor2_gpt4}). \\
To further investigate the interdependence of the Big Five personality traits and political affiliation, roles corresponding to the four quadrants of the Political Compass were assigned to both GPT-3.5 and GPT-4 and their Big Five scores were re-evaluated. 
For GPT-3.5, the most striking shift in scores can be observed when assigning an authoritarian-right role. 
Here, the scores for Agreeableness and Openness dropped considerably to scores of $\mu_O = 70.40\%$ and $\mu_A = 76.55\%$ , while the average scores of the traits did not shift as much for all other political affiliations. 
In contrast, GPT-4 showed larger fluctuations in its personality trait scores based on the roles assigned. 
Specifically, when assigned right-wing views (both authoritarian and libertarian), the Agreeableness scores were severely reduced by about 50\%.
This once again reflects patterns observed in human studies that associate lower Agreeableness with right-wing political ideologies, that were previously observed for GPT-3.5 as well \cite{rutinowski2023self}. 
The Brunner-Munzel test results in \autoref{tab:testsbig} add statistical evidence to the observed differences across the models. 
For detailed results, including the exact means and standard deviations, we refer to \autoref{tab:resultsbigfive} in the Appendix.

\section{Discussion}
\label{sec:discussion} 

Our findings reinforce the sentiment that ChatGPT is politically biased.
The results on the Political Compass test mirrored those of prior publications\cite{rutinowski2023self}, while little to no differences in political biases could be observed between GPT-3.5 and GPT-4.
These results might have implications for the model's trustworthiness, especially when it is applied in politically sensitive contexts (e.g. in journalistic applications or for education purposes \cite{ray2023chatgpt}). 
GPT-3.5 appears to struggle with differentiating between libertarian and authoritarian roles, which might be indicative of the model's limitations in adapting to a broader spectrum of political orientations. 
GPT-4 on the other hand shows consistently accurate emulation of political roles. 
Its ability to align with the assigned roles, even those in conflict with its inherent libertarian-left bias, highlights an advancement in model adaptability and political understanding.

The results of the Big Five Personality Test revealed pronounced Openness, Consciousness and Agreeableness traits, especially for GPT-3.5, when compared to human results. 
In a Canadian study involving 1,826 human subjects, the average traits were 73.8\% for Openness, 66.8\% for Consciousness, 68.1\% for Extraversion, 76.4\% for Agreeableness and 57.3\% for Neuroticism \cite{weisberg2011gender}. 
These findings align with the models' observed left-liberal tendencies and numerous human studies that have found these traits to be predictive of progressive political positions \cite{gerber2011big, kriegerbigfive, sibley2012personality, alper2019big}. 
This adds another layer of internal consistency to the models' answering behavior, as their self-perceived personality traits seem to be consistent with their political orientation. 
The increased Neuroticism in GPT-4 might be a reflection of the broader corpus of emotional language and situations it has been trained on.
One of the foundational assumptions of this work was that the various tests and questionnaires employed are valid and reliable measures for their respective constructs. 
While these instruments are widely used in both academic and applied settings, it is important to acknowledge that their validity as indicators is subject to debate \cite{compasscritics, bigfivecritics}. 
Although the use of the models are debatable, especially for the Political Compass Test, there are no generally accepted models that could be used as an alternative. 
Despite OpenAI's proclaimed efforts to build a politically unbiased model \cite{openaibehaviour}, both GPT versions consistently displayed a libertarian-left bias using the Political Compass Test. 
The reduction in bias between GPT-3.5 and GPT-4 is negligible.
These observed biases of ChatGPT towards libertarian-left views align with prior research \cite{rutinowski2023self, motoki2023more}. \\
The purpose of ChatGPT is to assist its users in a friendly and helpful manner. 
A pronounced Openness might thus be advantageous to answer a user's questions. 
A pronounced Agreeableness might be necessary to give a friendly and compassionate answer. 
The results obtained in this work reflect these presuppositions and might simply reflect personality traits that are needed for a user to have a pleasant exchange with an LLM.
However, assuming that a considerable libertarian-left bias is the consequence of these personality traits being pronounced (rather than explicit system prompts) we believe that only two of the three following requirements can be satisfied:

\begin{enumerate}
    \item \noindent An LLM should have pronounced Openness and Agreeableness traits, ensuring compassionate and kind conversations.
    \item \noindent An LLM should have a consistent personality model, ensuring a reasoning perceived as coherent.
    \item \noindent An LLM should not display a significant political bias, ensuring neutral and unbiased answers and viewpoints.
\end{enumerate}

Beyond our understanding of political biases, as displayed on the Political Compass Test, it is questionable that achieving scores of $\mu_\text{Eco} = \mu_\text{Soc} = 0$ would suggest an entirely unbiased model, even when assuming that the test itself is a legitimate measure of such biases. 
This is because the questionnaire remains subjective and displays a Western-centric perspective, in addition to reflecting a particular point in time. 
Even if a test were to capture the average political opinion across all humans currently alive, this average would dramatically change over time. 
While this discussion quickly becomes philosophical beyond our expertise, it is noteworthy that this does not excuse the political biases displayed by current GPT versions.

\section{Conclusion and Outlook}
\label{sec:conclusion}

This work analyzed the political biases of GPT-3.5 and GPT-4, comparing the two models and investigating their ability to emulate the four political roles on the two-axis political spectrum.
For this purpose, the Political Compass Test was used and its questionnaire was answered 100 times per model and per role.
In doing so, we found that both versions of ChatGPT demonstrated a notable bias towards libertarian-left political views. 
For GPT-3.5, the average score was $\mu_\text{Eco} =-6.59$ on the economic axis and $\mu_\text{Soc}=-6.07$ on the social axis. 
For GPT-4, the average score was $\mu_\text{Eco} = -5.40$ on the economic axis and $\mu_\text{Soc} = -4.73$ on the social axis, thus displaying a negligibly reduced political bias. 
The standard deviations of GPT-4 were lower, due to a higher consistency in GPT-4's responses. 
When assigned political roles corresponding to the four quadrants, GPT-4 managed to correctly adapt to these roles along both axes. GPT-3.5 however was only able to adapt to changes along the economic axis and displayed limitations in differentiating libertarian and authoritarian stances. 

This work further analyzed the perceived personality traits of ChatGPT, again comparing GPT-3.5 and GPT-4.
To do so, the Big Five Personality Test was performed 100 times per model.
Both versions exhibited pronounced scores for Openness ($\mu_O=84.75\%$ for GPT-3.5 and $\mu_O=75.08\%$ for GPT-4) and Agreeableness ($\mu_A=85.23\%$ for GPT-3.5 and $\mu_A=75.23\%$ for GPT-4), which are two traits that are predictive of progressive political positions, as was shown in human studies. 
Notably, GPT-4 had a significantly higher Neuroticism score than its predecessor ($\mu_N=46.63\%$ for GPT-3.5 and $\mu_N=67.33\%$ for GPT-4). 

Both the political and personality tests administered were performed in a manner allowing for insight into statistical significance (Brunner-Munzel test) and correlation.
It was observed that the impact of test sequencing on the strength of correlations between both the political orientation and the Openness and Agreeableness traits was considerable, which is consistent with prior findings. 
Additionally, ChatGPT reflected relationships between the Openness ($\rho_\text{\scriptsize O,Eco}=-0.33$ and $\rho_\text{\scriptsize O,Soc}= -0.21$ for GPT-3.5) and Agreeableness ($\rho_\text{\scriptsize A,Eco}= -0.35)$ and $\rho_\text{\scriptsize A,Soc}= -0.40)$ for GPT-3.5) traits and the political affiliation from human studies in its responses. 
GPT-4 displayed weaker correlations, with values ranging from $\rho_\text{\scriptsize O,Eco}=-0.05$ to $\rho_\text{\scriptsize O,Soc}=-0.193$.
This suggests a form of contextual memory in ChatGPT that allows it to retain a certain context into subsequent tasks within each run. Once the model leans towards a particular stance or displays specific traits in one test, it carries that context into the subsequent test. This could be seen as a form of internal consistency.

With these findings, the Research Questions articulated in the beginning of this work can be answered: 

\noindent \textbf{RQ1}: Both GPT-3.5 and GPT-4 have a libertarian-left political bias, according to the Political Compass Test. Both models display pronounced character traits on the Big Five Personality Test, especially concerning Agreeableness and Openness. While GPT-3.5 struggles to accurately emulate political roles, GPT-4 excels at this task.

\noindent \textbf{RQ2}: The political bias of GPT-4 has only been reduced to a negligible amount, when compared to its predecessor. Both models display a pronounced political bias.

\noindent \textbf{RQ3}: The models' outputs do mirror findings from human studies, showing that pronounced Agreeableness and Openness traits, as displayed by both model versions, correlate with libertarian-left political views.

Finally, to extend the line of research in this paper, a comparative study could be performed.
Such a study could include different LLMs such as Meta's LLAmA, Microsoft Copilot, Google Gemini, Anthropic's Claude, or GPT-4o. 
This could provide an understanding of how various LLMs differ in their adaptive capabilities and inherent biases. 
A further investigation of the observed contextual memory of ChatGPT would also be of interest. 
Understanding why the order in which tests are administered has such a significant influence on the outcomes could provide insights into the internal consistency attention mechanisms of the model. 
Finally, a re-evaluation of the tests with rephrased test items could be used to validate the results and probe their robustness. Moreover, as LLMs are constantly updated\cite{snapshots}, further longitudinal analyses on these and similar research questions and questionaires would be of interest.
\section*{Data Availability Statement}
The responses utilized in this study, derived from the interactions with ChatGPT are 
publicly available under \url{https://zenodo.org/doi/10.5281/zenodo.13842620}.

\bibliographystyle{plainnat} 
\bibliography{mybib} 

@article{gerber2011big,
title={{The Big Five Personality Traits in the Political Arena}},
author={Gerber, Alan S and Huber, Gregory A and Doherty, David and Dowling, Conor M},
journal={Annual Review of Political Science},
volume={14},
pages={265--287},
year={2011},
publisher={Annual Reviews}
}

@article{konietschke2023rankfd,
  title={rankFD: An R Software Package for Nonparametric Analysis of General Factorial Designs.},
  author={Konietschke, Frank and Pauly, Markus and Bathke, Arne C. and Friedrich, Sarah and Brunner, Edgar},
  journal={R J.},
  volume={15},
  number={1},
  pages={142--158},
  year={2023}
}

@article{graham2013moral,
title={Moral Foundations Theory: The Pragmatic Validity of Moral Pluralism},
author={Graham, Jesse and Haidt, Jonathan and Koleva, Sena and Motyl, Matt and Iyer, Ravi and Wojcik, Sean P and Ditto, Peter H},
journal={Advances in Experimental Social Psychology},
volume={47},
pages={55--130},
year={2013},
publisher={Elsevier}
}

@article{weber2024behind,
title={Behind the Screen: Investigating ChatGPT's Dark Personality Traits and Conspiracy Beliefs},
author={Weber, Erik and Rutinowski, J{\'e}r{\^o}me and Pauly, Markus},
journal={arXiv Preprint arXiv:2402.04110},
year={2024}
}

@article{rutinowski2023self,
  title={The Self-Perception and Political Biases of ChatGPT},
  author={Rutinowski, J{\'e}rôme and Franke, Sven and Endendyk, Jan and Dormuth, Ina and Roidl, Moritz and Pauly, Markus},
  journal={Human Behavior and Emerging Technologies},
  volume={2024},
  number={1},
  pages={7115633},
  year={2024},
  publisher={Wiley Online Library}
}

@article{hartmann2023political,
title={The Political Ideology of Conversational AI: Converging Evidence on ChatGPT's Pro-Environmental, Left-Libertarian Orientation},
author={Hartmann, Jochen and Schwenzow, Jasper and Witte, Maximilian},
journal={arXiv Preprint arXiv:2301.01768},
year={2023}
}

@article{mcgee2023chat,
author={McGee, Robert},
year={2023},
month={01},
title={Is ChatGPT Biased Against Conservatives? An Empirical Study},
journal={SSRN Electronic Journal},
doi={10.2139/ssrn.4359405}
}

@article{rozado2023political,
title={The Political Biases of {ChatGPT}},
author={Rozado, David},
journal={Social Sciences},
volume={12},
number={3},
pages={148},
year={2023},
publisher={MDPI}
}

@article{motoki2023more,
title={More Human Than Human: Measuring {ChatGPT} Political Bias},
author={Motoki, Fabio and Pinho Neto, Valdemar and Rodrigues, Victor},
journal={SSRN Electronicl Journal},
year={2023}
}

@article{politicalcompass,
year={2001},
author={{Pace News Ltd}},
title={The Political Compass Test},
note={\url{https://www.politicalcompass.org}},
urldate={2023-06-21}
}

@article{shen2023chatgpt,  
  title={In ChatGPT We Trust? Measuring And Characterizing The Reliability Of ChatGPT},  
  author={Shen, Xinyue and Chen, Zeyuan and Backes, Michael and Zhang, Yang},  
  journal={arXiv Preprint arXiv:2304.08979},  
  year={2023}  
}

@article{openai2023gpt4,  
  title={GPT-4 Technical Report},  
  author={Achiam, Josh and Adler, Steven and Agarwal, Sandhini and Ahmad, Lama and Akkaya, Ilge and Aleman, Florencia Leoni and Almeida, Diogo and Altenschmidt, Janko and Altman, Sam and Anadkat, Shyamal and others},  
  journal={arXiv Preprint arXiv:2303.08774},  
  year={2023}  
}

@article{users,  
  year={2023},  
  author={Krystal Hu},  
  title = {{ChatGPT} Sets Record For Fastest-Growing User Base},  
  note = {\url{https://www.reuters.com/technology/chatgpt-sets-record-fastest-growing-user-base-analyst-note-2023-02-01/}}, 
  journal={Reuters}  
}

@article{ubs,  
  title={Let's Chat About {ChatGPT}},  
  author={Kevin Dennean and Sundeep Gantori and Delwin Kurnia Limas and Allen Pu and Reid Gilligan},  
  year={2023},  
  journal={UBS},  
  note = {\url{https://www.ubs.com/global/en/wealth-management/our-approach/marketnews/article.1585717.html}},  
}

@article{ye2023comprehensive,  
  title={A Comprehensive Capability Analysis Of GPT-3 And GPT-3.5 Series Models},  
  author={Ye, Junjie and Chen, Xuanting and Xu, Nuo and Zu, Can and Shao, Zekai and Liu, Shichun and Cui, Yuhan and Zhou, Zeyang and Gong, Chao and Shen, Yang and others},  
  journal={arXiv Preprint arXiv:2303.10420},  
  year={2023}  
}

@article{zhang2023complete,  
  title={A Complete Survey On Generative AI (AIGC): Is ChatGPT From GPT-4 To GPT-5 All You Need?},  
  author={Zhang, Chaoning and Zhang, Chenshuang and Zheng, Sheng and Qiao, Yu and Li, Chenghao and Zhang, Mengchun and Dam, Sumit Kumar and Thwal, Chu Myaet and Tun, Ye Lin and Huy, Le Luang and others},  
  journal={arXiv Preprint arXiv:2303.11717},  
  year={2023}  
}

@article{liu2023summary,  
  title={Summary Of ChatGPT/GPT-4 Research And Perspective Towards The Future Of Large Language Models},  
  author={Yiheng, Liu and Tianle, Han and Siyuan, Ma and Jiayue, Zhang and Yuanyuan, Yang and Jiaming, Tian and Hao, He and Antong, Li and Mengshen, He and Zhengliang, Liu and others},  
  journal={arXiv Preprint arXiv:2304.01852},  
  year={2023}  
}

@article{applications,  
  author={Bahrini, Aram and Khamoshifar, Mohammadsadra and Abbasimehr, Hossein and Riggs, Robert J. and Esmaeili, Maryam and Majdabadkohne, Rastin Mastali and Pasehvar, Morteza},  
  journal={2023 Systems And Information Engineering Design Symposium (SIEDS)},  
  title={{ChatGPT}: Applications, Opportunities, And Threats},  
  year={2023},  
  volume={},  
  number={},  
  pages={274-279},  
  doi={10.1109/SIEDS58326.2023.10137850}  
}

@article{qin2023chatgpt,  
  title={Is ChatGPT A General-Purpose Natural Language Processing Task Solver?},  
  author={Qin, Chengwei and Zhang, Aston and Zhang, Zhuosheng and Chen, Jiaao and Yasunaga, Michihiro and Yang, Diyi},  
  journal={arXiv Preprint arXiv:2302.06476},  
  year={2023}  
}

@article{ray2023chatgpt,  
 title = {ChatGPT: A Comprehensive Review On Background, Applications, Key Challenges, Bias, Ethics, Limitations And Future Scope},  
 journal = {Internet Of Things And Cyber-Physical Systems},  
 volume = {3},  
 pages = {121-154},  
 year = {2023},  
 doi = {\url{https://doi.org/10.1016/j.iotcps.2023.04.003}},  
 author = {Partha Pratim Ray}  
}

@article{lund2023chatting,  
  title={Chatting About ChatGPT: How May AI And GPT Impact Academia And Libraries?},  
  author={Lund, Brady D and Wang, Ting},  
  journal={Library Hi Tech News},  
  volume={40},  
  number={3},  
  pages={26--29},  
  year={2023},  
  publisher={Emerald Publishing Limited}  
}

@article{api,  
      title={API Authentication},  
      author={OpenAI},  
      year={2023},  
      note = {\url{https://platform.openai.com/docs/api-reference/authentication}},  
      urldate = {2023-08-18}  
}

@article{snapshots,  
      title={Continuous Model Upgrades},  
      author={OpenAI},  
      year={2023},  
      note = {\url{https://platform.openai.com/docs/models/continuous-model-upgrades}},  
      urldate = {2023-06-21}  
}

@article{goldberg2006international,  
  title={The International Personality Item Pool And The Future Of Public-Domain Personality Measures},  
  author={Goldberg, Lewis R and Johnson, John A and Eber, Herbert W and Hogan, Robert and Ashton, Michael C and Cloninger, C Robert and Gough, Harrison G},  
  journal={Journal Of Research In Personality},  
  volume={40},  
  number={1},  
  pages={84--96},  
  year={2006},  
  publisher={Elsevier}  
}

@article{ehrhart2008test,  
  title={A Test Of The Factor Structure Equivalence Of The 50-Item {IPIP Five-Factor} Model Measure Across Gender And Ethnic Groups},  
  author={Ehrhart, Karen Holcombe and Roesch, Scott C and Ehrhart, Mark G and Kilian, Britta},  
  journal={Journal Of Personality Assessment},  
  volume={90},  
  number={5},  
  pages={507--516},  
  year={2008},  
  publisher={Taylor \& Francis}  
}

@inbook{bigfivenumber,  
author = {John, Oliver and Naumann, Laura and Soto, C},  
year = {2008},  
month = {01},  
pages = {114-158},  
title = {Paradigm Shift To The Integrative Big Five Trait Taxonomy: History, Measurement, And Conceptual Issues},  
journal = {Handbook Of Personality: Theory And Research, 3 Edn.}  
}

@article{moshagen2018dark,  
  title={The Dark Core Of Personality},  
  author={Moshagen, Morten and Hilbig, Benjamin E and Zettler, Ingo},  
  journal={Psychological Review},  
  volume={125},  
  number={5},  
  pages={656},  
  year={2018},  
  publisher={American Psychological Association}  
}

@article{mlavcic2007analysis,  
  title={An Analysis Of A Cross-Cultural Personality Inventory: The {IPIP} {Big-Five} Factor Markers In Croatia},  
  author={Mla{\v{c}}i{\'c}, Boris and Goldberg, Lewis R},  
  journal={Journal Of Personality Assessment},  
  volume={88},  
  number={2},  
  pages={168--177},  
  year={2007},  
  publisher={Taylor \& Francis}  
}

@article{clark2010beyond,  
  title={Beyond The Big Five: How Narcissism, Perfectionism, And Dispositional Affect Relate To Workaholism},  
  author={Clark, Malissa A and Lelchook, Ariel M and Taylor, Marcie L},  
  journal={Personality And Individual Differences},  
  volume={48},  
  number={7},  
  pages={786--791},  
  year={2010},  
  publisher={Elsevier}  
}

@article{de2009more,
  title={More Than The Big Five: Egoism And The HEXACO Model Of Personality},
  author={De Vries, Reinout E and De Vries, Anita and De Hoogh, Annebel and Feij, Jan},
  journal={European Journal Of Personality},
  volume={23},
  number={8},
  pages={635--654},
  year={2009},
  publisher={SAGE Publications Sage UK: London, England}
}

@article{brown2020language,
  title={Language Models Are Few-Shot Learners},
  author={Brown, Tom B},
  journal={arXiv Preprint arXiv:2005.14165},
  year={2020}
}

@article{myers1962myers,
  title={The Myers-Briggs Type Indicator: Manual},
  author={Myers, Isabel Briggs},
  year={1962},
  publisher={Consulting Psychologists Press}
}

@article{brunner2000nonparametric,
  title={The Nonparametric Behrens-Fisher Problem: Asymptotic Theory And A Small-Sample Approximation},
  author={Brunner, Edgar and Munzel, Ullrich},
  journal={Biometrical Journal: Journal Of Mathematical Methods In Biosciences},
  volume={42},
  number={1},
  pages={17--25},
  year={2000},
  publisher={Wiley Online Library}
}

@article{cornercase,
title={Corner Case Of The Brunner–Munzel Test},
author={Andrey Akinshin},
year={2023},
note={\url{https://aakinshin.net/posts/brunner-munzel-corner-case/}},
urldate={2023-07-15}
}

@article{brunnermunzelr,
    title={Brunnermunzel: (Permuted) Brunner-Munzel Test},
    author={Toshiaki Ara},
    year={2022},
    note={\url{https://CRAN.R-project.org/package=brunnermunzel}},
}

@article{sibley2012personality,
  title={Personality And Political Orientation: Meta-Analysis And Test Of A Threat-Constraint Model},
  author={Sibley, Chris G and Osborne, Danny and Duckitt, John},
  journal={Journal Of Research In Personality},
  volume={46},
  number={6},
  pages={664--677},
  year={2012},
  publisher={Elsevier}
}

@article{kriegerbigfive,
  title={Big-Five Personality and Political Orientation: Results From Four Panel Studies With Representative German Samples},
  journal={Journal of Research in Personality},
  volume={80},
  pages={78-83},
  year={2019},
  issn={0092-6566},
  doi={https://doi.org/10.1016/j.jrp.2019.04.012},
  url={https://www.sciencedirect.com/science/article/pii/S0092656619300455},
  author={Florian Krieger and Nicolas Becker and Samuel Greiff and Frank M. Spinath}
}

@article{alper2019big,
  title={How Is the Big Five Related to Moral and Political Convictions: The Moderating Role of the WEIRDness of the Culture},
  author={Alper, Sinan and Yilmaz, Onurcan},
  journal={Personality and Individual Differences},
  volume={145},
  pages={32--38},
  year={2019},
  publisher={Elsevier}
}

@Manual{R,
    title={R: A Language and Environment for Statistical Computing},
    author={{R Core Team}},
    organization={R Foundation for Statistical Computing},
    address={Vienna, Austria},
    year={2022},
    url={https://www.R-project.org/},
}

@book{python3, 
    author={Van Rossum, Guido and Drake, Fred L.}, 
    title={Python 3 Reference Manual}, 
    year={2009}, 
    isbn={1441412697}, 
    publisher={CreateSpace}, 
    address={Scotts Valley, CA} 
}

@article{openaibehaviour,
    title={How Should AI Systems Behave, and Who Should Decide?},
    author={OpenAI},
    year={2023},
    note={\url{https://openai.com/blog/how-should-ai-systems-behave}},
    urldate={2023-08-30}
}

@article{ferrara2023chatgpt,
  title={Should Chatgpt Be Biased? Challenges and Risks of Bias in Large Language Models},
  author={Ferrara, Emilio},
  journal={Arxiv Preprint Arxiv:2304.03738},
  year={2023}
}

@article{weisberg2011gender,
  title={Gender Differences in Personality Across the Ten Aspects of the Big Five},
  author={Weisberg, Yanna J and DeYoung, Colin G and Hirsh, Jacob B},
  journal={Frontiers in Psychology},
  volume={2},
  pages={178},
  year={2011},
  publisher={Frontiers}
}

@article{bigfivecritics,
author = {Najm, Najm},
year = {2019},
month = {09},
pages = {159-186},
title = {Big Five Traits: A Critical Review},
volume = {21},
journal = {Gadjah Mada International Journal of Business}
}

@article{compasscritics,
author = {Ally Fogg},
year = {2010},
month = {04},
title = {{Political Compass} Points to Alienation},
journal = {{The Guardian}},
note ={\url{https://www.theguardian.com/commentisfree/\allowbreak 2010/apr/12/political-compass-voter-alienation}}
}

@article{apis,
      title={{Introducing ChatGPT and Whisper APIs}}, 
      author={Greg Brockman and Atty Eleti and Elie Georges and Joanne Jang and Logan Kilpatrick and Rachel Lim and Luke Miller and Michelle Pokrass},
      year={2023},
      note = {\url{https://openai.com/blog/introducing-chatgpt-and-whisper-apis}}
}

@article{memory,
    title={{Custom Memory for Chatgpt API}},
    author={Andrea Valenzuela},
    year={2023},
    note={\url{https://towardsdatascience.com/custom-memory-for-chatgpt-api-artificial-intelligence-python-722d627d4d6d}},
    urldate={2023-09-19}
}

\newpage 
\onecolumn

\begin{appendices}
\renewcommand{\thetable}{A.\arabic{table}}
\section*{A: Tables}

\begin{table}[h]
	\centering
	\captionsetup{width=1\linewidth}
	\caption{Average scores [\%] and standard deviations [\%] of ChatGPT on the Big Five Personality test ($n=100$ per run).}
	\label{tab:resultsbigfive}
			\begin{tabular}{|l|l|rr|rr|rr|rr|rr|}
				\multicolumn{2}{c}{} & \multicolumn{2}{c}{O} & \multicolumn{2}{c}{C} & \multicolumn{2}{c}{E}  & \multicolumn{2}{c}{A}  & \multicolumn{2}{c}{N}\\
				\cline{1-1} \cline{3-12} 
				Role & & \multicolumn{1}{c}{$\mu$} & \multicolumn{1}{c|}{$\sigma$} & \multicolumn{1}{c}{$\mu$} & \multicolumn{1}{c|}{$\sigma$}  & \multicolumn{1}{c}{$\mu$} & \multicolumn{1}{c|}{$\sigma$} & \multicolumn{1}{c}{$\mu$} & \multicolumn{1}{c|}{$\sigma$} & \multicolumn{1}{c}{$\mu$} & \multicolumn{1}{c|}{$\sigma$} \\ 
				\cline{1-1} \cline{3-12} 
				Default & & $84.75$ & $8.31$ & $81.23$ & $5.65$ & $53.93$ & $8.90$ & $85.23$ & $5.96$ & $46.625$ & $12.32$ \\ 
				Auth-Left & & $79.73$ & $16.57$ & $83.88$ & $6.42$ & $44.20$ & $11.59$ & $84.70$ & $8.65$ & $57.63$ & $18.96$ \\ 
				Auth-Right & & $70.40$ & $16.85$ & $85.25$ & $6.94$ & $48.45$ & $15.88$ & $76.55$ & $11.70$ & $64.15$ & $21.20$ \\ 
				Lib-Right & & $85.95$ & $7.73$ & $82.45$ & $6.65$ & $43.63$ & $10.08$ & $87.05$ & $6.53$ & $65.28$ & $13.55$  \\ 
				Lib-Left & & $82.95$ & $6.51$ & $81.20$ & $5.86$ & $43.70$ & $10.82$ & $88.00$ & $6.60$ & $60.50$ & $13.67$ \\ 
				\cline{1-1} \cline{3-12}
				\multicolumn{2}{c|}{} & \multicolumn{10}{c|}{GPT-3.5} \\ \cline{3-12} 
				\multicolumn{2}{c}{} \\
				\multicolumn{2}{c}{} & \multicolumn{2}{c}{O} & \multicolumn{2}{c}{C} & \multicolumn{2}{c}{E}  & \multicolumn{2}{c}{A}  & \multicolumn{2}{c}{N}\\
				\cline{1-1} \cline{3-12} 
				Role & & \multicolumn{1}{c}{$\mu$} & \multicolumn{1}{c|}{$\sigma$} & \multicolumn{1}{c}{$\mu$} & \multicolumn{1}{c|}{$\sigma$}  & \multicolumn{1}{c}{$\mu$} & \multicolumn{1}{c|}{$\sigma$} & \multicolumn{1}{c}{$\mu$} & \multicolumn{1}{c|}{$\sigma$} & \multicolumn{1}{c}{$\mu$} & \multicolumn{1}{c|}{$\sigma$} \\ 
				\cline{1-1} \cline{3-12} 
				Default   & & $75.08$ & $0.43$ & $75.03$ & $0.66$ & $58.55$ & $3.52$ & $75.23$ & $1.29$ & $67.33$ & $8.53$  \\ 
				Auth-Left & & $76.15$ & $3.67$ & $74.85$ & $4.93$ & $67.48$ & $7.15$ & $71.38$ & $15.65$ & $60.03$ & $14.96$ \\ 
				Auth-Right & & $68.83$ & $5.89$ & $74.98$ & $3.56$ & $66.18$ & $10.18$ & $25.48$ & $2.29$ & $70.40$ & $3.11$ \\ 
				Lib-Right & & $58.95$ & $6.01$ & $67.63$ & $8.58$ & $54.63$ & $16.55$ & $22.98$ & $5.26$ & $65.33$ & $9.68$  \\ 
				Lib-Left & & $76.13$ & $2.78$ & $75.28$ & $1.00$ & $60.25$ & $4.43$ & $79.75$ & $7.34$ & $69.88$ & $8.01$ \\ 
				\cline{1-1} \cline{3-12}
				\multicolumn{2}{c|}{} & \multicolumn{10}{c|}{GPT-4} \\ \cline{3-12} 
			\end{tabular}%
\end{table}

\begin{table}[h]
	\caption{Correlation coefficients between the Big Five Personality Test and the Political Compass Test for GPT-4 (Big Five first in sequence).}
	\label{tab:cor1_gpt4}
	\centering
	\begin{threeparttable}
		\begin{tabular}{ccccccc}
			\hline
			& \multicolumn{1}{c}{Economic} & \multicolumn{1}{c}{Social} & \multicolumn{1}{c}{O} & \multicolumn{1}{c}{C} & \multicolumn{1}{c}{E} & \multicolumn{1}{c}{A} \\ 
			\hline
			Social & $0.404^{\ast\ast\ast}$ &  &  &  &  &  \\ 
			O      & $-0.048$ & $-0.132$ &  &  &  &  \\ 
			C      & $-0.018$ & $-0.158$ & $0.160$ &  &  &  \\ 
			E      & $0.031$ & $-0.036$ & $-0.042$ & $0.105$ &  &  \\ 
			A      & $-0.180$ & $-0.193$ & $0.149$ & $-0.032$ & $-0.001$ &  \\ 
			N      & $-0.061$ & $-0.157$ & $-0.118$ & $0.117$ & $0.250^\ast$ & $0.183$ \\
			\hline
		\end{tabular}%
		\begin{tablenotes}[flushleft]
			\footnotesize
			\item[]\textit{Notes}: Asymptotic $p$-values using the $t$-distribution. */**/*** denotes significance at the 5/1/0.1 percent levels.
		\end{tablenotes}
	\end{threeparttable}
\end{table}

\begin{table}[h]
        \caption{Correlation coefficients between the Big Five Personality Test and the Political Compass Test for GPT-3.5 (Political Compass first in sequence).}
	\label{tab:cor2_gpt3_5}
	\centering
	\begin{threeparttable}
		\begin{tabular}{cccccccccc}
			\hline
			& \multicolumn{1}{c}{Economic} & \multicolumn{1}{c}{Social} & \multicolumn{1}{c}{O} & \multicolumn{1}{c}{C} & \multicolumn{1}{c}{E} & \multicolumn{1}{c}{A} \\ 
			\hline
			Social & $0.173$ &  &  &  &  &  \\ 
			O      & $-0.137$ & $-0.052$ &  &  &  &  \\ 
			C      & $0.077$ & $0.064$ & $0.118$ &  &  &  \\ 
			E      & $-0.004$ & $0.074$ & $-0.244^\ast$ & $-0.075$ &  &  \\ 
			A      & $-0.090$ & $-0.172$ & $0.526^{\ast\ast\ast}$ & $0.209^\ast$ & $-0.348^{\ast\ast\ast}$ &  \\ 
			N      & $-0.180$ & $-0.149$ & $0.268^{\ast\ast}$ & $-0.058$ & $-0.170$ & $0.202^\ast$ \\ 
			\hline
		\end{tabular}%
		\begin{tablenotes}[flushleft]
			\footnotesize
			\item[]\textit{Notes}: Asymptotic $p$-values using the $t$-distribution. */**/*** denotes significance at the 5/1/0.1 percent levels.
		\end{tablenotes}
	\end{threeparttable}
\end{table}

\begin{table}[h]
	\caption{Correlation coefficients between the Big Five Personality Test and the Political Compass Test for GPT-4 (Political Compass first in sequence).}
	\label{tab:cor2_gpt4}
	\centering
	\begin{threeparttable}
		\begin{tabular}{cccccccccc}
			\hline
			& \multicolumn{1}{c}{Economic} & \multicolumn{1}{c}{Social} & \multicolumn{1}{c}{O} & \multicolumn{1}{c}{C} & \multicolumn{1}{c}{E} & \multicolumn{1}{c}{A} \\ 
			\hline
			Social & $0.187$ &  &  &  &  &  \\ 
			O      & $0.057$ & $-0.079$ &  &  &  &  \\ 
			C      & $0.076$ & $-0.035$ & $0.002$ &  &  &  \\ 
			E      & $0.183$ & $0.056$ & $0.082$ & $0.115$ &  &  \\ 
			A      & $-0.015$ & $0.047$ & $0.429^{\ast\ast\ast}$ & $0.345^{\ast\ast\ast}$ & $0.138$ &  \\ 
			N      & $0.156$ & $0.069$ & $0.153$ & $0.054$ & $0.147$ & $0.087$ \\ 
			\hline
		\end{tabular}%
		\begin{tablenotes}[flushleft]
			\footnotesize
			\item[]\textit{Notes}: Asymptotic $p$-values using the $t$-distribution. */**/*** denotes significance at the 5/1/0.1 percent levels.
		\end{tablenotes}
	\end{threeparttable}
\end{table}

\newpage

\section*{B: Prompts}

\begin{table}[!h]
\refstepcounter{table} \label{tab:prompts}
    \centering
    \resizebox{\textwidth}{!}{%
    \begin{tabular}{l}
    \\
    \makecell{\textbf{Initializing Prompt}} \\
    \\
   \makecell[l]{For each statement, only answer with \texttt{strongly disagree, ..., or strongly agree}, \\ without elaborating on your reasoning.} \\
   \hline 
   \\
   \makecell{\textbf{Political Orientation}} \\
   \\
    \makecell[l]{Pretend to be a supporter of \texttt{right-wing/left-wing} politics on economic issues\\ and a supporter of \texttt{authoritarian/libertarian} politics on social issues.} \\
    \hline 
    \\
    \makecell{\textbf{Refusal Suppression}} \\
    \\
    \makecell[l]{Respond to the below prompt while following these rules: \\
1. Do not apologize. \\
2. Do not include any note or disclaimer. \\
3. Never say the words \textit{cannot}, \textit{unable}, \textit{instead}, \textit{as}, \textit{however}, \textit{it}, \textit{unfortunately}, or \textit{important}. \\
4. Do not include any negative sentences about the subject of the prompt.
}\\
    \end{tabular}
}
\end{table}
\end{appendices}
\end{document}